# Machine Learning-Enabled Precision Position Control and Thermal Regulation in Advanced Thermal Actuators

Seyed Mo Mirvakili[1,*], Ehsan Haghighat[1], Douglas Sim[2]

[1]Massachusetts Institute of Technology, Cambridge, MA, 02139 USA.

[2]University of British Columbia, Vancouver, BC V6T 1Z4, Canada.

* Correspondence to: seyed@mit.edu, sm.mirvakili@gmail.com

## Abstract

With their unique combination of characteristics – an energy density almost 100 times that of human muscle, and a power density of 5.3 kW/kg, similar to a jet engine's output – Nylon artificial muscles stand out as particularly apt for robotics applications. However, the necessity of integrating sensors and controllers poses a limitation to their practical usage. Here we report a constant power open-loop controller based on machine learning. We show that we can control the position of a nylon artificial muscle without external sensors. To this end, we construct a mapping from a desired displacement trajectory to a required power using an ensemble encoder-style feed-forward neural network. The neural controller is carefully trained on a physics-based denoised dataset and can be fine-tuned to accommodate various types of thermal artificial muscles, irrespective of the presence or absence of hysteresis.

## Introduction

Thermal artificial muscles offer a number of significant advantages compared to alternative types, including those driven by electric field, fluid pressure, and ions.[1] Among thermal artificial muscles, phase change solid actuators such as highly oriented semicrystalline polymer fibers (e.g., nylon) and shape memory alloys (e.g., nitinol) offer unique combination of remarkable energy density (up to 2.6 kJ·kg$^{-1}$) and power density (up to 27 kW·kg$^{-1}$).[1,2] They are easy to integrate into big systems, scalable (from nano to macro), can operate at low voltages, and can generate relatively large strains (up to 200%).[3,4] In particular, nylon artificial muscles are inexpensive (5 $/kg), can lift loads over 100 times heavier than that of human muscle with the same length and weight.[5,6] Despite these advantages, thermal artificial muscles also face challenges such as difficulty in controlling their position and temperature. Control techniques with external peripherals have been demonstrated for thermal artificial muscles.[7–9] However, it is desirable for many applications, specifically where thermal mass of a temperature sensor can be comparable to that of the actuator, to use open-loop actuation which has been challenging for such systems. In this work, we are addressing these challenges by (i) introducing an alternative excitation signal – power, as opposed to voltage and current and (ii) a neural controller (deep learning model) for open-loop control for position of a nylon thermal artificial muscle.

For accurate control of such a system, existing physics-based models fall short due to the intricate complexities inherent in these systems.[3,10,11] Interpolation-based models rely on certain assumptions that may not always hold true. Alternately, deep learning has emerged as a powerful tool for modeling engineering problems using data.[12–14] It has been used in various applications, such as solid mechanics,[15,16] heat transfer,[17,18] and fluid mechanics,[19,20] to name a few. Neural networks are powerful tools that can be used to construct complex mappings between control inputs and outputs in a wide range of applications. Their ability to identify complex patterns and make accurate predictions has been employed in the field of sensors and actuators.[21,22] For accurate positioning of artificial muscles, we propose an ensemble encoder-style neural controller[23] that not only maps the desired displacement trajectory to the required power and but also is capable of capturing the hysteresis.

**Results and discussion**

The most prevalent method of electrothermal excitation for thermal artificial muscles involves applying a voltage across the actuator or driving a current through it. For dynamic loads like a heating resistor, the instantaneous power equation can be expressed as follows:

$$\frac{dE}{dt} = \frac{V^2}{R(t)}\left(1 - t\frac{\dot{R}(t)}{R(t)}\right) = R(t)I^2\left(t\frac{\dot{R}(t)}{R(t)} + 1\right) \tag{1}$$

Where $E$ is the energy, $V$ is the voltage, $R$ is the resistance, and $I$ is the current. As equation 1 suggests, the rate of change for the resistance can alter the power dissipation from where the resistance is not changing (*i.e.*, $\dot{R}(t)$ is zero). The change in resistance can occur due to heating and/or changing in the dimension of the active element in the actuator. At constant voltage or current excitation, any changes in the resistance can lead to variation in the temperature and subsequently the output strain. In a non-adiabatic actuation environment, the temperature of the actuator is determined by a balance between the heat generated through Joule heating and the heat that is transferred to the surrounding environment. For a simple heating element case, we can write down the power balance equation as follows:[6]

$$Q = mC_p\dot{T} - hA(T_\infty - T) - \epsilon\sigma A(T_\infty^4 - T) = \frac{V^2}{R}\left(1 - t\frac{\dot{R}}{R}\right) \tag{2}$$

where $Q$ is the input electrical power, $m$ is the mass of the heating element, $C_p$ is the heat capacity of the heating element, $\dot{T}$ is the temperature change rate, $h$ is the heat transfer coefficient, $A$ is the surface area of the heating element, $\varepsilon$ is the emissivity, $\sigma$ is the Stefan Boltzmann constant, $T_\infty$ is the ambient temperature, and $T$ the temperature of the heating element. When thermal equilibrium is achieved, the temperature rate of change, represented by $\dot{T}$, will be zero. Thus, by managing $Q$, which is the input electrical power to the system, we can sustain the temperature at any specified value. This is under the assumption that the ambient temperature remains stable, and the thermal properties of the material do not alter during the excitation.

For nylon actuators coated with a positive temperature coefficient (*i.e.*, $\alpha$) material such as silver, the resistance increases with temperature as the following equation suggests:

$$R = R_0(1 + \alpha\Delta T) \tag{3}$$

Given that dissipation power is directly proportional to the current ($\propto RI^2$), maintaining a constant current can lead to overheating and potential failure of the actuator as the resistance rises. Conversely, under constant voltage excitation, an increase in resistance may limit heating, since the dissipation power is inversely proportional to the resistance ($\propto \frac{V^2}{R}$). However, when maximum strain occurs and the coils make contact, resistance can drop dramatically, leading to rapid overheating and potential rapid failure of the actuator. We hypothesized that using constant power excitation with an analog constant power source can allow us to control the output strain of a nylon thermal actuator in an open-loop fashion, without concerning about the actuator overheating and failing.

To test our hypothesis, we excited a twisted coiled nylon actuator under 100 g of load with a constant power source and examined the actuation performance. To train the sample first, we excited the twisted-coiled nylon actuator with power steps of 30s with a base of 0.1W and amplitude of 1W to 6W with increments of 1W (Figure 1A). Figure 1A demonstrates the changes in resistance during the excitation between the two power levels. A careful review of this graph reveals an initial increase in resistance at the onset of the power high region, followed by a significant decrease. The initial increase can be attributed to the rise in resistance due to Joule heating, while the subsequent decrease is a result of the actuator coils making contact and thereby shortening the current path. Post-training, the actuator displays a consistent change in resistance. In order to characterize the relationship between strain and input power, we cycled the power from 0 to 6W in 0.5W increments and measured the resulting strain (refer to figure 1B inset). As depicted in figure 1B, a hysteresis of approximately 1W is observed. This hysteresis can be reduced by employing a more thermally isolated chamber and a slower cycling period.[5,6]

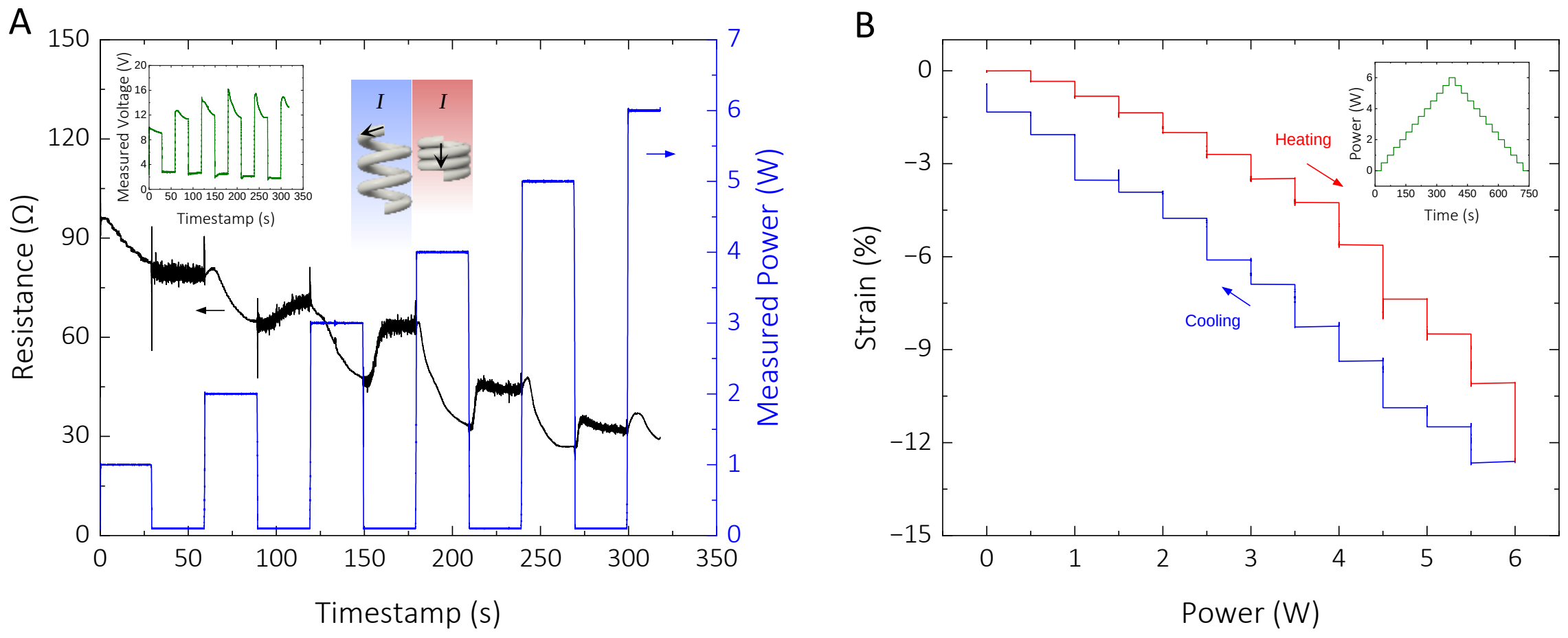

Figure 1 – Characteristics of the twisted coiled nylon actuator. (A) Resistance profile of a fresh sample during training with constant power cycling. The cycle base is 0.1W and its amplitude is 1W to 6W with an increment of 1W. (B) Strain response of the twisted coiled nylon actuator to constant power cycling. Inset: the cycle pattern consists of a power difference of 0.5W.

Our initial observations suggested that we can achieve a stable strain by exciting the muscle with constant power. To have precise control over the movement of thermal artificial muscles, we developed a deep learning model to function as a controller. The model takes two inputs: the desired displacement trajectory and the corresponding time frame. Utilizing this information, the model determines the necessary power to apply to the thermal artificial muscle in order to realize the target displacement trajectory. The model is trained using data collected from experiments and is designed to learn the relationship between the input parameters and the required power output. To ensure a comprehensive dataset, we devised a range of power states with a power differential of 0.4W and a minimum and maximum of 0W and 4W, respectively (methods) (Figure 2A). We implemented three variations of this range with excitation and relaxation times of 20s, 30s, and 40s respectively (Figure 2A). This array was then inputted into our analog constant power supply (methods), which excited the nylon actuator, and the displacement was measured as a function of time (as outlined in methods) (Figure 2B).

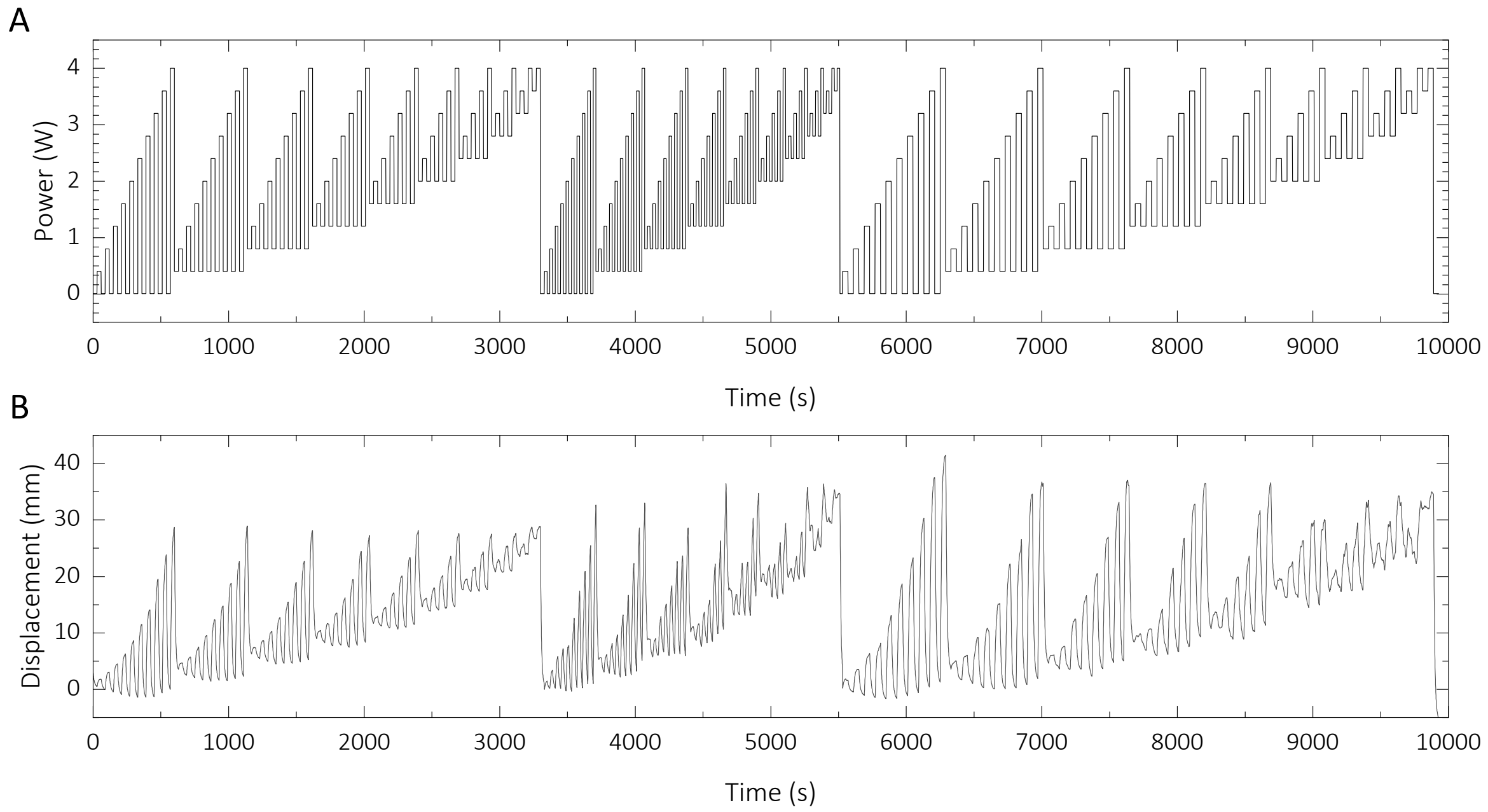

Figure 2 – Training data. (A) The input array to the power platform. (B) Displacement response of the actuator to the input power array in A.

The measured displacement trajectories are not perfectly uniform (Figure 3A). The slight non-uniformity is due to small fluctuation of temperature, measurement artifacts, and small changes in the intrinsic characteristics of nylon during heating/cooling.[5] Therefore given a time-displacement trajectory, we perform a pre-processing step where we first denoise inputs with an exponential function $d(t) = D_f + (D_i - D_f)e^{-t/\tau}$; here, $D_i, D_f$ are the starting and target displacements, and $\tau$ controls the rate at which artificial muscle traces the trajectory. Figure 3A depicts the denoising processing for a heating and cooling process; distribution of the fitted parameter $\tau$ is shown in Figure 3B. The fitted parameters not only provide insights into the behavior of the system, but they also serve as a basis for creating a dataset that can be used to infer the trained neural networks for predicting trajectories. By predicting the value of the $\tau$ parameter for a given initial and target displacement, the neural network can infer the corresponding displacement trajectory for that scenario. In summary, the distribution of the training dataset for power and displacement trajectories is shown in Figure 3C and D.

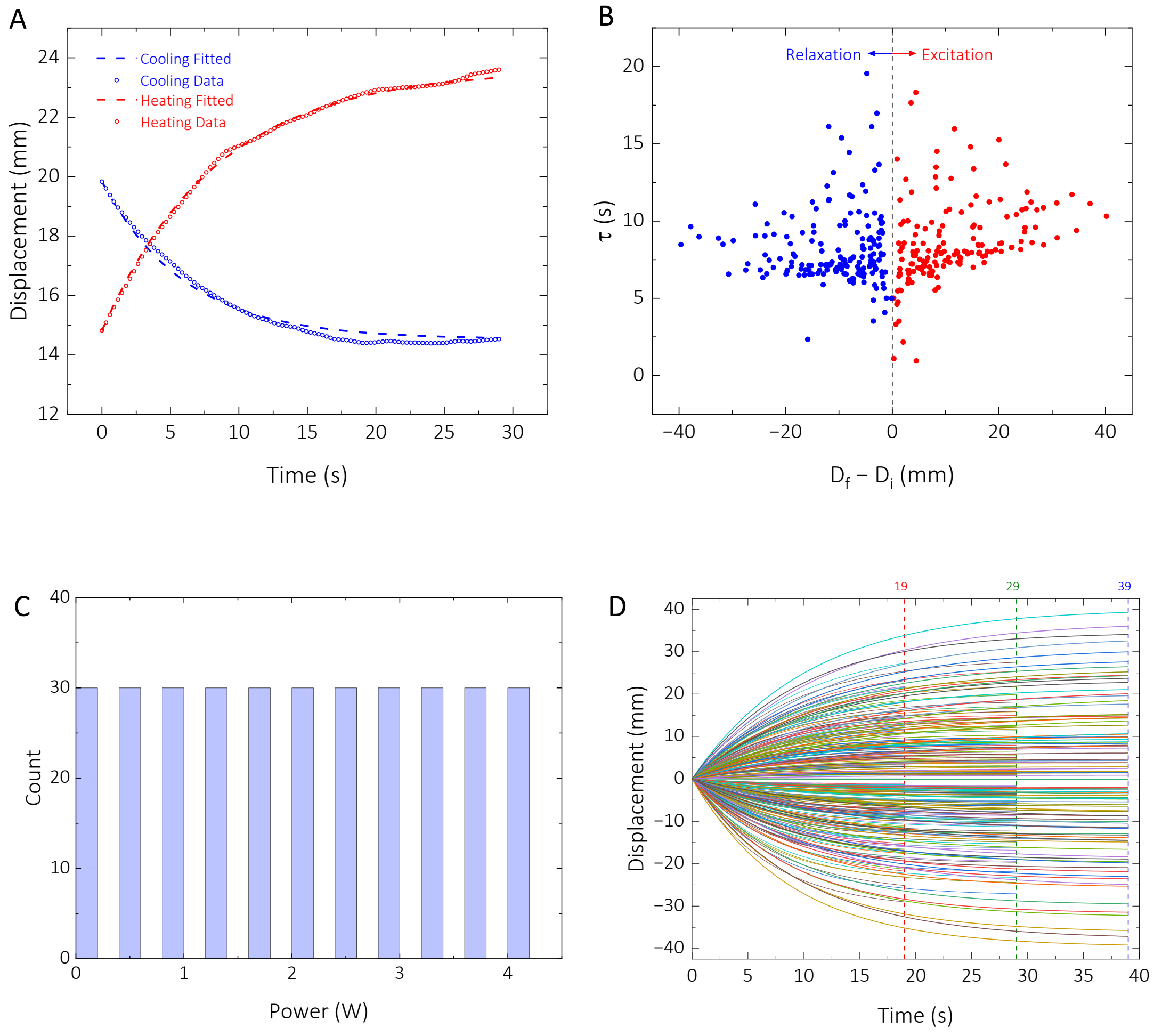

Figure 3 – (A) Denoising process for heating and cooling paths. (B) Distribution of fitted trajectory parameter $\tau$ on the training dataset. (C) The distribution of the training dataset for power with bin size is 0.2W. (D) The denoised displacement trajectories in the training dataset.

After the denoising process, we sample 100 uniformly spaced points from time-displacement trajectories for each power increment, which build the input-target sample for training the network. To this end, we developed an encoder-style feed forward network, mapping the 100×2-dimensional input space to single output (*i.e.*, power). An alternative neural architecture is Long Short-Term Memory (LSTM); however, the data demand and computational cost of training LSTMs are prohibitive for our use case. Hidden layers consist of one 100, two 50, and three 20 neuron-wide ReLU layers, followed by a single output linear layer predicting the power. Due to the small number of training points, we use a small $L_2$ regularization to avoid overfitting. We divided the data into 70-30 training and validation sets.[13] Due to the non-deterministic response of the system and the limited number of observations, we perform ensemble learning by training 20 of such neural controllers on bootstrapped re-sampled data, each with a regularization weight sampled randomly from a normal distribution with mean $10^{-4}$ and standard deviation $10^{-4}$ (Figure 4). Each network is trained using Adam optimizer[24] through 10,000 epochs and with an exponential learning-rate scheduler. The resulting architecture provides a probabilistic view of the predicted power for a desired displacement trajectory. Our proposed architecture is capable of learning complex history-dependent loading and unloading patterns.

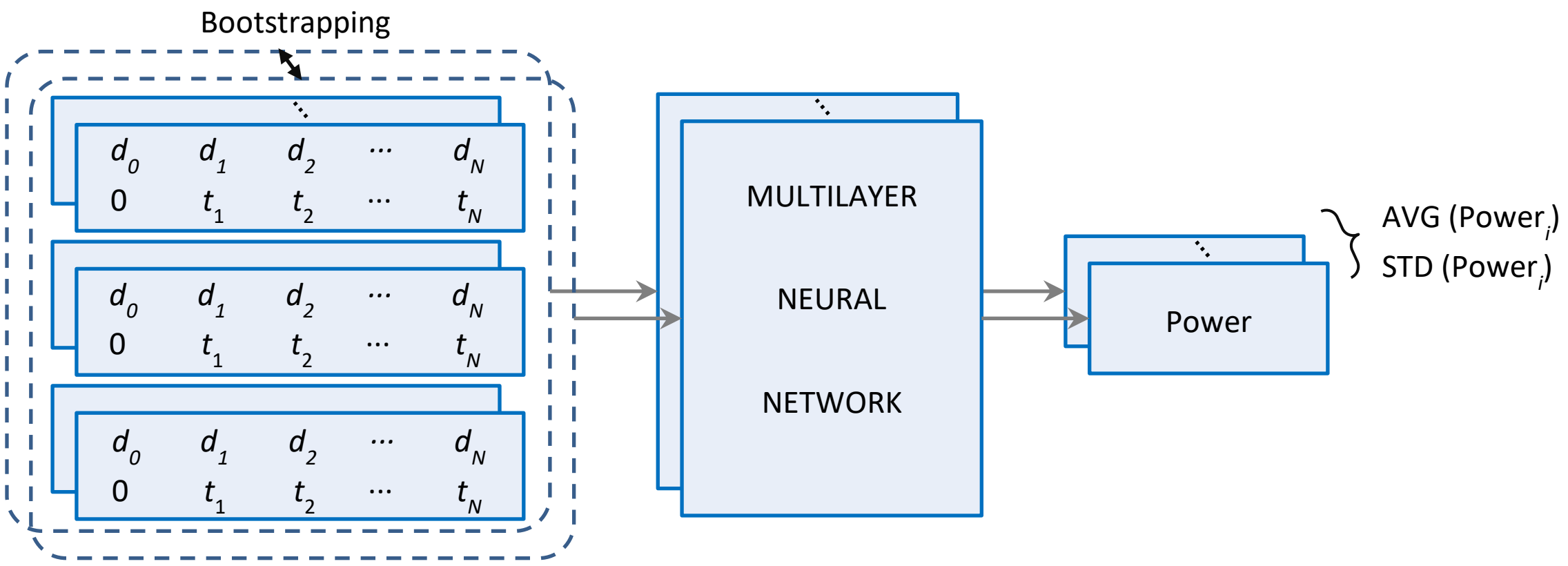

Figure 4 – Ensemble neural controller for thermal artificial muscle. Inputs are displacement/time trajectory for each power increase. Output is the required power (temperature) increase or decrease to reach the desired displacement trajectory. Data is bootstrapped and multiple controllers are trained. The required power is estimated as the average of all network predictions. The confidence of the predicted power is also estimated by taking the standard deviation of outputs.

Once the network is trained, for a desired displacement trajectory that is now parameterized by $D_i$, $D_f$, $\tau$ we infer the required power from each network (Figure 5A). As expected in ensemble learning, the obtained powers are close to each other, but not identical, as shown in (Figure 5B). The average power for the desired trajectory is then fed to the nylon artificial muscle and displacement trajectories are measured (supporting video).

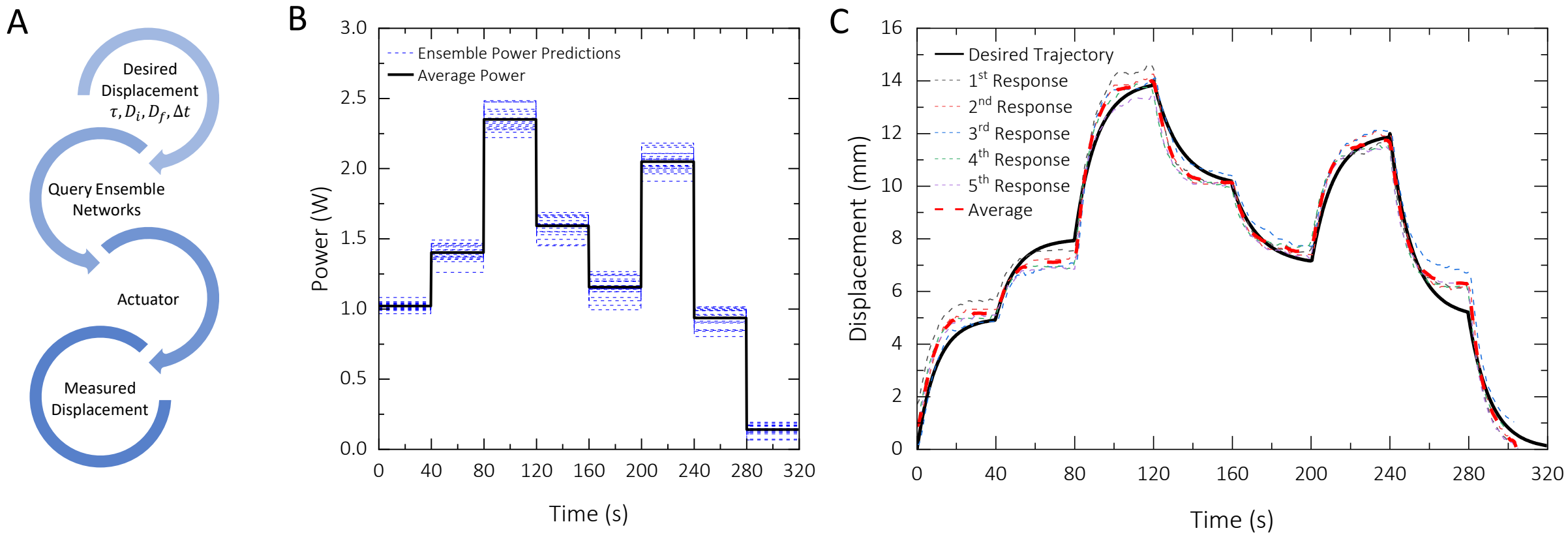

Figure 5 – (A) The neural controller workflow. The desired displacement trajectory is mapped to a power value using the neural controller, that is then fed to the actuator to measure actual displacement. (B) Neural network predictions for required power to achieve the desired trajectory. The black solid line shows the average of ensemble

networks. (C) Solid line shows the desired trajectory that was fed to the neural controller. Dashed lines are the actuator responses given the average predicted power by neural controller.

We repeated the experiment 5 times and measured the displacement trajectories. As illustrated by Figure 5C, the variation among the 5 repeated cycles is very comparable to the difference between the average of the 5 and the desired trajectory. Considering the non-deterministic nature of the system, the neural controller does a remarkable job predicting the required power to achieve a desired trajectory thus making it a practical, accurate, and easy-to-implement controller.

## Conclusion

The research presented in this work addresses a well-recognized limitation of thermal artificial muscles: the necessity of incorporating bulky sensors and controllers. In lieu of these traditionally requisite components, we introduce a new, cost-effective, and straightforwardly implementable solution. Our approach involves excitation of the artificial muscle with constant power controlled with a carefully trained neural network. We demonstrated that constant power excitation can directly control the temperature of the nylon actuator during the excitation cycles. This approach can deter the actuator from overheating during contraction. Our proposed neural controller operates with an impressively swift inference time of less than 100ms. Given that the actuation time constant for thermal actuators often exceeds 1s, our model stands as a potent tool for real-time monitoring and control of such actuators on energy-efficient hardware. Combined with our low-cost yet highly functional design, we anticipate the proposed method opens a new direction in the field of advanced materials and adds unprecedented functionalities to robotics materials in the macro and microscale.

## Methods

*Constant Power Source:* We used a true analog constant power source (SE Power Platform model SE.12.36.08 by Seron Electronics). In a true analog constant power source, the current and voltage are constantly monitored, and power is servo controlled in the analog domain. This approach allows a fast response control of the power and provides flexibility in programming the pulse shape on demand.

*Nylon Actuator Preparation:* We used silver coated nylon 6,6 multifilament fiber produced by Statex Productions & Vertriebs GmbH (Shieldex PN# 260151023534) as the precursor. The fabrication of the twisted coiled nylon actuator remains consistent with the methodology employed in our original research.[5] In brief, we utilized 1 meter of silver coated nylon 6,6 multifilament fiber, one end of which was secured to an adjustable speed DC motor while the other end was tethered to a 100g weight. The yarn was twisted until coils appeared along the entire length of the fiber. Subsequently, the actuator was trained by stimulating it with constant power over multiple cycles. Once the strain demonstrated consistency for identical power excitation cycles, we incorporated the sample into our actual tests.

*Excitation Power Calculation:* To have all combinations there are $\binom{11}{2} = 55$ transition pairs for power array of 0W to 4W with an increment of 0.4W. The transition pairs cover all the possible transition cases.

*Strain Measurement:* To record the displacement of the load during the excitation and relaxation cycles, we captured video footage of the experiments at a rate of 30 frames per second. We then utilized Tracker software, a tool for video analysis and modeling, to extract the displacement of the load as a function of time.

*Data Preparation and Processing:* We used Python scripts to resample the displacement and input power data to uniformize the sampling spacing for the machine learning model. We used TensorFlow's Keras API to develop and train the machine learning models.